\documentclass[sigconf]{acmart}
\usepackage{xspace}
\usepackage{enumitem}
\usepackage{balance}
\newcommand{\sys}{ELF-Gym\xspace}

\AtBeginDocument{%
  }

\copyrightyear{2024} 
\acmYear{2024} 
\setcopyright{acmlicensed}\acmConference[CIKM '24]{Proceedings of the 33rd ACM International Conference on Information and Knowledge Management}{October 21--25, 2024}{Boise, ID, USA}
\acmBooktitle{Proceedings of the 33rd ACM International Conference on Information and Knowledge Management (CIKM '24), October 21--25, 2024, Boise, ID, USA}
\acmDOI{10.1145/3627673.3679153}
\acmISBN{979-8-4007-0436-9/24/10}

\begin{document}

\title{\sys: Evaluating Large Language Models Generated Features for Tabular Prediction}

\author{Yanlin Zhang}
\affiliation{%
  \institution{Fudan University}
  \city{Shanghai}
  \country{China}
}
\email{21210720043@m.fudan.edu.cn}

\author{Ning Li}
\affiliation{%
  \institution{Shanghai Jiao Tong University}
  \city{Shanghai}
  \country{China}
}
\email{lining01@sjtu.edu.cn}

\author{Quan Gan}
\affiliation{%
  \institution{Amazon Shanghai AI Lab}
  \city{Shanghai}
  \country{China}
}
\email{quagan@amazon.com}

\author{Weinan Zhang}
\affiliation{%
  \institution{Shanghai Jiao Tong University}
  \city{Shanghai}
  \country{China}
}
\email{wnzhang@sjtu.edu.cn}

\author{David Wipf}
\affiliation{%
  \institution{Amazon Shanghai AI Lab}
  \city{Shanghai}
  \country{China}
}
\email{daviwipf@amazon.com}

\author{Minjie Wang}
\affiliation{%
  \institution{Amazon Shanghai AI Lab}
  \city{Shanghai}
  \country{China}
}
\email{minjiw@amazon.com}

% \renewcommand{\shortauthors}{Zhang et al.}
\renewcommand{\shortauthors}{Yanlin Zhang et al.}
%% No italics, no superscripts
%% Use footnote or author note to identify equal contribution and/or contact author info

\begin{abstract}
%\quan{Generated by GPT-4.  Do not treat it seriously.} In the realm of data science competitions, particularly those involving tabular data, feature engineering remains a critical component for enhancing predictive performance. While many attempts that automates feature engineering with large language models (LLMs) exist, few attempt to systematically compare the quality of generated features against human experts. This paper therefore introduces a novel benchmark and evaluation protocol designed for such comparison. Our benchmark comprises the "gold" feature engineering practices collected from the top-three solutions based on the discussion forums of 8 Kaggle competitions. Moreover, we established a protocol to evaluate the efficacy of LLM feature engineering, by expressing feature engineering algorithms from human and LLMs using Python code snippets, and counting how many human snippets are equivalent to any of the LLM's snippets. Our results reveal that advanced LLMs, such as GPT-4, can replicate only XXX\% of the "gold" features identified in human solutions. Furthermore, we observed that LLMs often overlook critical aspects such as high-order aggregations and temporal causality across rows. These findings highlight the current limitations of LLMs in automating the feature engineering process and a significant gap compared to human experts. Our benchmark serves as a foundation for future research aimed at improving the feature engineering capabilities of LLMs, ultimately advancing the automation of data science tasks.

Crafting effective features is a crucial yet labor-intensive and domain-specific task within machine learning pipelines. Fortunately, recent advancements in Large Language Models (LLMs) have shown promise in automating various data science tasks, including feature engineering. But despite this potential, evaluations thus far are primarily based on the end performance of a complete ML pipeline, providing limited insight into precisely how LLMs behave relative to human experts in feature engineering. To address this gap, we propose \sys , a framework for \underline{E}valuating \underline{L}LM-generated \underline{F}eatures. We curated a new dataset from historical Kaggle competitions, including 251 ``golden'' features used by top-performing teams. \sys then quantitatively evaluates LLM-generated features by measuring their impact on downstream model performance as well as their alignment with expert-crafted features through semantic and functional similarity assessments. This approach provides a more comprehensive evaluation of disparities between LLMs and human experts, while offering valuable insights into specific areas where LLMs may have room for improvement.  For example, using \sys we empirically demonstrate that, in the best-case scenario, LLMs can semantically capture approximately 56\% of the golden features, but at the more demanding implementation level this overlap drops to 13\%.   Moreover, in other cases LLMs may fail completely, particularly on datasets that require complex features, indicating broad potential pathways for improvement.

\end{abstract}

% Using \sys, we conducted a study to highlight key differences and performance gaps between LLM-generated and human-engineered features.


%    <concept>
%        <concept_id>10010147.10010257</concept_id>
%        <concept_desc>Computing methodologies~Machine learning</concept_desc>
%        <concept_significance>300</concept_significance>
%        </concept>
%    <concept>
%        <concept_id>10010147.10010178.10010179.10010182</concept_id>
%        <concept_desc>Computing methodologies~Natural language generation</concept_desc>
%        <concept_significance>300</concept_significance>
%        </concept>
%  </ccs2012>


% \ccsdesc[500]{Computing methodologies~Machine learning}
%\ccsdesc[300]{Computing methodologies~Machine learning}
%\ccsdesc[300]{Computing methodologies~Natural language generation}


\newcommand{\quan}[1]{{\textcolor{red}{[Quan: #1]}}}
\newcommand{\mj}[1]{{\textcolor{blue}{[MJ: #1]}}}
\newcommand{\david}[1]{{\textcolor{purple}{[David: #1]}}}
% \newcommand{\quan}[1]{}
% \newcommand{\mj}[1]{}
% \newcommand{\david}[1]{}

\maketitle

\section{Introduction}
Feature engineering is a crucial step in the machine learning pipeline, transforming raw data into meaningful features that improve model performance and interpretability. Effective feature engineering can significantly enhance the predictive power of models, making it a vital component in various data-driven applications. This importance is particularly evident in competitive data science environments like Kaggle~\cite{kaggle}, where top-performing models often rely heavily on sophisticated feature engineering techniques. For instance, in one interview \footnote{\url{https://medium.com/kaggle-blog/grupo-bimbo-inventory-demand-winners-interview-clustifier-alex-andrey-1e3b6cec8a20}}, the winners of the Grupo Bimbo Inventory Prediction competition reported that 95\% of their time was on feature engineering while only 5\% was on modeling.

%Despite its importance, traditional feature engineering poses several challenges. It is often labor-intensive, requiring extensive domain knowledge and significant time investment. Numerous automated feature engineering tools have emerged to address these challenges by streamlining the feature creation process. AutoFeat~\cite{horn2020autofeat}, for example, automates feature selection and generation using statistical methods and heuristics. OpenFE~\cite{zhang2022openfe} focuses on generating features from relational data, leveraging graph-based algorithms. SAFE~\cite{shi2020safe} employs a supervised approach to feature engineering, optimizing feature sets based on their predictive power. These tools have made strides in reducing the manual effort involved in feature engineering, but they still fall short in fully capturing the semantic richness of data.

Despite its importance, traditional feature engineering is labor-intensive and requires extensive domain knowledge. Automated tools like AutoFeat~\cite{horn2020autofeat}, OpenFE~\cite{zhang2022openfe}, SAFE~\cite{shi2020safe}, and Deep Feature Synthesis (DFS)~\cite{kanter2015deep} have emerged to streamline this process. AutoFeat automates feature selection and generation using statistical methods and heuristics but suffers from high feature generation costs. OpenFE and SAFE mitigate these costs 
by optimizing the feature selection phase using feedback from model evaluation.
%through efficient feature selection \david{This sentence seems circular, as in: ``These methods mitigate feature selection costs by efficient feature selection.''  Can something more informative be said about these methods?}.
DFS extends feature engineering to multi-table scenarios by utilizing data relationships to generate features. Despite their effectiveness in reducing manual effort, these tools often fall short in leveraging the domain knowledge that human experts typically rely on for crafting relevant features.

The advent of Large Language Models (LLMs) such as GPTs~\cite{achiam2023gpt} has opened new possibilities for automating various data science tasks. LLMs have demonstrated remarkable capabilities in natural language understanding~\cite{min2023recent, du2023shortcut}, text generation~\cite{yuan2022wordcraft, lu2023bounding}, summarization~\cite{goyal2022news, zhang2024benchmarking, basyal2023text}, and even code synthesis~\cite{austin2021program, chen2021evaluating, zhang2023planning}. Their ability to process and generate human-like text makes them particularly well-suited for tasks that require semantic understanding and contextual reasoning. Of particular relevance to data science, LLMs have shown potential in automating tasks such as data cleaning, feature generation, and model selection.

For example, recent work such as CAAFE~\cite{hollmann2024large}, DS-Agent~\cite{guo2024ds} and FeatLLM~\cite{han2024large} have explored the application of LLMs to feature engineering. In brief, CAAFE leverages LLMs to generate additional features based on dataset descriptions, iteratively improving model performance through semantic understanding. Meanwhile DS-Agent employs a case-based reasoning approach, combining LLMs with expert knowledge from Kaggle competitions to automate the entire data science workflow. Finally, FeatLLM utilizes LLMs to engineer binary features through rule generation and rule parsing, significantly improving down-stream tabular prediction tasks.
Despite the potential, their actual evaluations thus far are primarily based on the end performance of a complete machine learning pipeline, providing limited insight into the reason behind the distinction between the solutions built by LLMs and human experts especially with respect to feature engineering. As LLM-based data science is increasingly becoming an active research area, this calls for more transparent and insightful evaluation tools and benchmarks to effectively assess and improve these systems. %\mj{cite the ICML24 paper \cite{huang2023benchmarking}; the method doesn't seem to include complex feature patterns like groupby.}\david{If this paper is cited, need to clearly differentiate from this submission.}
% \david{What are key metrics for quantifying different feature engineering strategies besides downstream performance?  For example, are there some metrics related to interpretability?}

To make strides in this direction, we proposed \sys, a framework for \underline{E}valuating \underline{L}LM-generated \underline{F}eatures in machine learning pipelines. We curate a new dataset specifically designed for evaluating LLMs in feature engineering tasks, using historical data from Kaggle competitions. This dataset includes 251 so-called "golden" features used by top-5 teams in 8 Kaggle competitions. Using those features as references, we then develop toolkits to quantitatively evaluate the quality of features generated by LLMs. In addition to measuring the impact of LLM-generated features on downstream model performance, \sys also evaluates their alignment with expert-crafted features by assessing semantic and functional similarity. This provides a more direct measurement of the gap between LLMs and data science experts, offering valuable insights into specific areas where LLMs need improvement.
%the quality and impact of LLM-engineered features. Our evaluation toolkit measures the quality of LLM-based features versus expert-engineered features via different angles. First, we 
%via three criteria: (1), (2) whether they compute the same result, \david{Are the first two intended to be binary evaluations?  Present wording suggests so.} and (3) how they contribute to the end model performance. \david{Would be helpful to provide brief motivation here for why these three criteria were chosen.}

Leveraging the proposed toolkit, we conducted a pioneer study involving multiple popular LLMs
% (e.g., Mixtral~\cite{jiang2024mixtral}, GPT-4o~\cite{achiam2023gpt}, LLaMA3~\cite{llama3modelcard} and Claude~\cite{anthropic2024claude})
\cite{jiang2024mixtral,achiam2023gpt,llama3modelcard,anthropic2024claude}
to highlight the key differences and performance gaps between LLM-generated and human-engineered features. In doing so we focused on two main research questions:%\david{These research questions and the findings below could potentially be more precisely aligned with the three evaluation metrics mentioned above.}
\begin{itemize}[leftmargin=1em,topsep=0pt]
    \item \textbf{RQ1}: \textit{Can LLMs discover golden features by reasoning from data descriptions and schemas?}
    \item \textbf{RQ2}: \textit{What are common golden feature patterns that LLMs excel at generating, or conversely, struggle to produce?}
\end{itemize}
For RQ1, we find that LLMs can capture at best approximately 56\% of the "golden" features by description, but only 13\% at best when coming to implementation, and may fail completely on datasets that require complex features. For RQ2, we found that while LLMs are capable of generating features based on simpler patterns such as feature interactions or simple group-by operations, they struggle to discover patterns involving multi-variable functions, custom aggregations and complex table joins.
These new findings again highlight the necessity of a standardized benchmark and robust evaluation tools, and how the proposed new evaluation metrics against human experts can derive more insights into the area for LLMs to improve.

\begin{figure}
\vspace*{-1.5em}
    \centering
    \includegraphics[width=0.8\columnwidth]{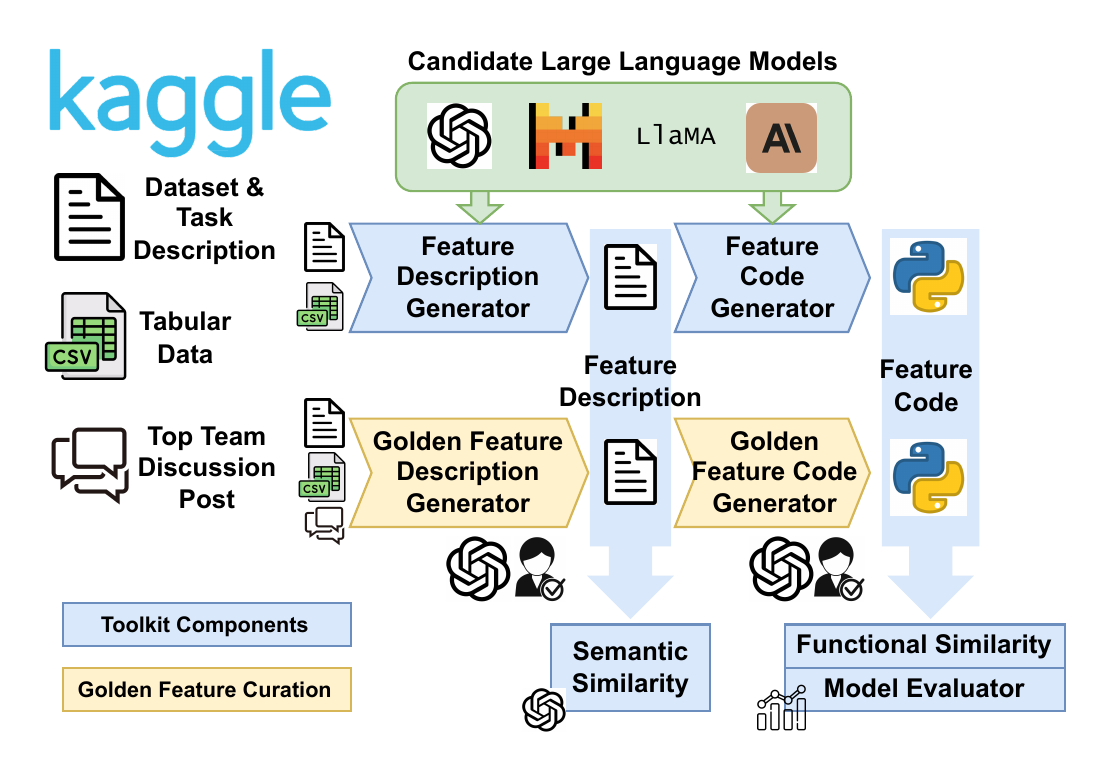}
    \caption{\sys evaluation framework overview. Blue components are part of the evaluation pipeline while the yellow ones are only triggered during golden feature curation.} 
    \label{fig:pipeline}
    \vspace*{-1.5em}
\end{figure}

%\mj{Describe Figure~\ref{fig:pipeline}  and connect to subsections.}
\section{\sys Design}

\subsection{Dataset and Golden Feature Collection}

Kaggle is a widely used platform in the data science community, providing a rich source of datasets for developing and testing LLM-based methods. It hosts numerous competitions where data science experts compete to build the best models, often employing sophisticated feature engineering techniques. This makes Kaggle an ideal source for comparing features discovered by experts with those generated by LLMs.

However, despite competition hosts inviting or enforcing winner teams to share their code and insights in discussion posts, the code is often hard to consume as it was written in R, C++, Python, etc., well before scientific computing packages like Pandas \cite{reback2020pandas} became popular.  As per the insights, the discussion posts are typically written casually and are often entangled with unrelated discussions, making it difficult to extract relevant information directly.
% However, the challenge lies in the fact that competition participants rarely release their code \david{Is this possible to quantify?} \quan{This depends; some competitions enforce that at least the top-rankers must open source the code.}\david{Sure, but is it possible to say something like, over a random sample of x competitions, only y\% of entries released the code?}. Fortunately, competition hosts often invite top teams to share their insights and key features on the discussion forum. These posts, while valuable, are typically written casually and are often entangled with unrelated discussions, making it difficult to extract relevant information directly.

To address this, we employ an LLM-assisted strategy to convert discussion posts into features that can be calculated from actual data. Similar to~\cite{han2024large}, we adopt a two-step approach to ensure reliability. For each Kaggle dataset, we provide GPT-4o with the dataset description, including table and column names, column labels, and the intended task (i.e., the column to predict), and prompt GPT-4o to extract feature descriptions from the discussion posts of the top-5 teams in a specific format. Once we have the feature descriptions, we further prompt GPT-4o to generate code that computes actual feature given the original table. GPT-4o can generate correct code approximately 80\% of the time,
%\mj{check number} \quan{I found we changed 50 code blocks in total.  Dividing this by 251 gives us 80\%.  I might be wrong though as we didn't really keep track of how many actual features we fixed, and the 50 code blocks may overlap so the actual ratio might be higher.},
significantly reducing the time of manual checking. We then manually sanitize the remaining cases where the code fails to run.

To validate the effectiveness of these extracted golden features, we test if these features can enhance model performance. This is done by executing the generated code, adding the resulting features to the original tabular data, and using AutoGluon~\cite{erickson2020autogluon} for prediction. We compare the results with predictions made using only the raw features.  With this approach, we successfully curated 251 golden features from 8 Kaggle datasets with details in Table~\ref{tab:data-stats}.

%We selected eight Kaggle competitions.

%Four competitions permit usage of the data for academic publications: Avito Context Ad Clicks (\textbf{Avito}) \cite{avito-context-ad-clicks}, Instacart ..., Outbrain Click Prediction (\textbf{Outbrain}) \cite{outbrain-click-prediction}, and IEEE-CIS Fraud Detection (\textbf{IEEE-CIS}) \cite{ieee-fraud-detection}.

%The other four competitions forbid the usage of data outside competition, so \emph{we only collect their dataset descriptions}, which are accessible even without agreeing to the competition rules.  They are: AirBnB New User Bookings (\textbf{AirBnB}) \cite{airbnb-recruiting-new-user-bookings}, Facebook Recruiting IV: Human or Robot (\textbf{Facebook}) \cite{facebook-recruiting-iv-human-or-bot}, West Nile Virus Prediction (\textbf{WestNile}) \cite{predict-west-nile-virus}, and TalkingData Mobile User Demographics (\textbf{TalkData}) \cite{talkingdata-mobile-user-demographics}.  We hypothesize that feature engineering is still possible even by looking at the description alone, without accessing the actual data.  We verify our hypothesis in Section~\ref{sec:results}.

\begin{table}[t]
  \small
  \caption{Dataset statistics.  Asterisks(*) indicate that the licenses prohibit use of data for academic publications.}
  \label{tab:data-stats}
  \begin{tabular}{ccc}
    \toprule
    \textbf{Name} & \textbf{Task Description} & \textbf{\# Golden Feats} \\
    \midrule
    \textbf{Avito \cite{avito-context-ad-clicks}} & Click-through-rate Prediction & 30 \\
    \textbf{Instacart \cite{instacart-market-basket-analysis}} & Product Prediction & 24 \\
    \textbf{Outbrain \cite{outbrain-click-prediction}} & Click-through-rate Prediction & 30 \\
    \textbf{IEEE-CIS \cite{ieee-fraud-detection}} & Fraud Detection & 27 \\
    \textbf{AirBnB* \cite{airbnb-recruiting-new-user-bookings}} & Destination Prediction & 52 \\
    \textbf{Facebook* \cite{facebook-recruiting-iv-human-or-bot}} & Bot Detection & 32 \\
    \textbf{WestNile* \cite{predict-west-nile-virus}} & Virus Prediction & 25 \\
    \textbf{TalkingData* \cite{talkingdata-mobile-user-demographics}} & Demographic Prediction & 31 \\
  \bottomrule
\end{tabular}
\vspace*{-3mm}
\end{table}

%Among the eight competitions, we collected the solution description from top-five teams.  Each solution description includes a number of \emph{gold} features, which they handcraft as functions of existing columns in the data.  The number of gold features for each dataset is shown in Table~???\ref{tab:stats}.  The gold features are in various forms ranging from Python or R code to natural language text or even simply column names.  To enable rigorous and automated comparison of LLM's generation against the gold features, we paraphrase the gold features into Pandas \cite{reback2020pandas} code.  
%\quan{TODO: should we say that they are generated by GPT (and maybe manually tweaked by human)?}

\subsection{Generating Features Using LLMs} \label{sec:gen_LLM_features}

Similar to how golden features are curated, the LLM-generated features also take two steps (Figure \ref{fig:pipeline}). First, the candidate LLM receives the dataset description and prediction target and is asked to generate a list of feature descriptions. The feature descriptions together with the dataset and task descriptions are then passed to the second stage for code generation. Our toolkit provides two components called \textbf{Feature Description Generator} and \textbf{Feature Code Generator} where users can plug in their LLMs for evaluation. Both the feature descriptions and code generated by LLMs are compared with the golden feature descriptions and code for evaluation, as we will explain next.

%the evaluation framework (Figure ???\ref{fig:pipeline}) comprises two main components: the \textbf{Feature Description Generator}, which takes the dataset's description and table contents to generate feature descriptions in a fixed format, and the \textbf{Feature Code Generator}, which generates code based on these descriptions. Both components allow users to customize the LLMs in use, making it easier to setup comparison across models.

%We propose to measure LLM's feature engineering capability via \emph{recall}, i.e. the ratio between the number of gold features equivalently generated by LLM and the total number of gold features.  The equivalence between an LLM feature and a gold feature is determined probabilistically: we generate KK synthetic input tables, pass them to the Pandas code of the gold feature and the Pandas code generated by the LLM, and compare the output.  If all KK outputs are equal, then we say that the LLM feature is equivalent to the gold feature.  \quan{TODO: add details of how the synthetic tables are generated.}

%\quan{TODO: describe the UX of our package.}

%Similar to the curation of golden features, the evaluation framework (Figure ??????\ref{fig:pipeline}) comprises two main components: the \textbf{Feature Description Generator} takes the dataset's description and table contents to generate feature descriptions in a fixed format, while the \textbf{Feature Code Generator} then generates code based on these descriptions. Both components allow for LLM customization to ensure fair comparisons.\david{How does LLM customization relate to fairness?}

\subsection{Evaluation Protocol}

To measure the alignment between LLM-engineered features and golden features, \sys employs a recall metric. Given two feature sets, $F_{LLM}$ and $F_{golden}$, the recall metric uses a measurement function $M$ that returns a binary flag (1 or 0), indicating whether two features are similar. The recall metric is then defined as the proportion of golden features for which the $M$ returns 1 when compared to features in $F_{LLM}$:

\vspace{-0.3cm}
$$
\text{Recall}(F_{LLM}, F_{golden}) = \frac{\sum_{f \in F_{golden}} \max_{f' \in F_{LLM}} M(f, f')}{|F_{golden}|}
$$
\vspace{-0.3cm}

\sys uses two measurement functions $M$. $M_{sem}$ measures the semantic similarity of two feature descriptions by prompting a GPT-4o to assess and score the similarity. $M_{func}$ checks if two features are functionally equivalent by comparing the outputs of feature functions applied to the input data. This process can be costly due to the slow performance of LLM-generated codes on large datasets, so we employ strategies such as representative down-sampling to mitigate issues like false positives. For instance, down-sampling must be done carefully to ensure that key patterns in the data are preserved, especially for features involving aggregations by specific IDs. The use of two $M$ functions is important because, for example, if a feature is hit by $M_{sem}$ but not $M_{func}$, it indicates that the LLM can identify relevant features but needs to improve its code generation capability for feature engineering tasks. We remark that while a precision metric can be defined in a similar fashion, we omit it here since the usefulness of a feature crafted by LLMs, but not overlapping with human experts, requires downstream model evaluation.

In addition to these alignment measures, \sys also evaluates the impact of LLM-generated features on downstream model performance. This involves incorporating the generated features into the original dataset and using them to train a model, then comparing the model's performance to that of a model trained on raw features. This step provides a direct measure of the utility of the LLM-engineered features. \sys also supports scenarios where table data is unavailable due to restricted licenses, allowing for the evaluation of LLM feature generation capabilities based solely on data descriptions.

%\sys employs three metrics to evaluate LLM-engineered features. Firstly, we measure whether the generated feature descriptions are semantically similar to the golden descriptions, using an LLM to assess and score the similarity. Secondly, we check if the generated feature code is functionally equivalent to golden features by comparing the outputs of feature functions applied to the input data. This process can be costly due to the slow performance of LLM-generated codes on large datasets, so we employ strategies such as representative down-sampling to mitigate issues such as false positives. Finally, we also evaluate how the downstream model performs with the generated features compared to the raw features, providing a direct measure of their utility.

%For both description similarity and functionality equivalence, we calculate \mj{TODO: recall? hit?}. \sys also supports scenarios where table data is unavailable due to restricted licenses, which allows for testing LLM feature generation capabilities based solely on data descriptions.

\section{Initial Results}
\label{sec:results}

We select four LLMs to evaluate: GPT-4o \cite{achiam2023gpt}, Claude 3 Sonnet \cite{anthropic2024claude}, LLaMA3-70B-Instruct \cite{llama3modelcard}, and Mixtral-8x7B-Instruct \cite{jiang2024mixtral}; by design though, \sys can also be readily adapted to handle other LLMs (see Section \ref{sec:gen_LLM_features}).  We invoke GPT-4o with OpenAI's official SDK and the rest three with Amazon Bedrock\footnote{https://aws.amazon.com/bedrock/}.  For feature description generation, we explicitly prompt the LLMs to generate as many features as possible.  For code generation, we gave each LLM three chances to write runnable code, each time feeding in the error message from Python interpreter as the next round of conversation.  Our implementation is available at \url{https://github.com/Lilyzhangyanlin/ELF-Gym}. 

% \mj{Experiment setup: what models, what datasets, other details like we let LLMs generate features until natural stops.}

\subsection{RQ1: Can LLMs Discover Golden Features?}

\begin{figure}
    \centering
    \includegraphics[width=0.8\linewidth]{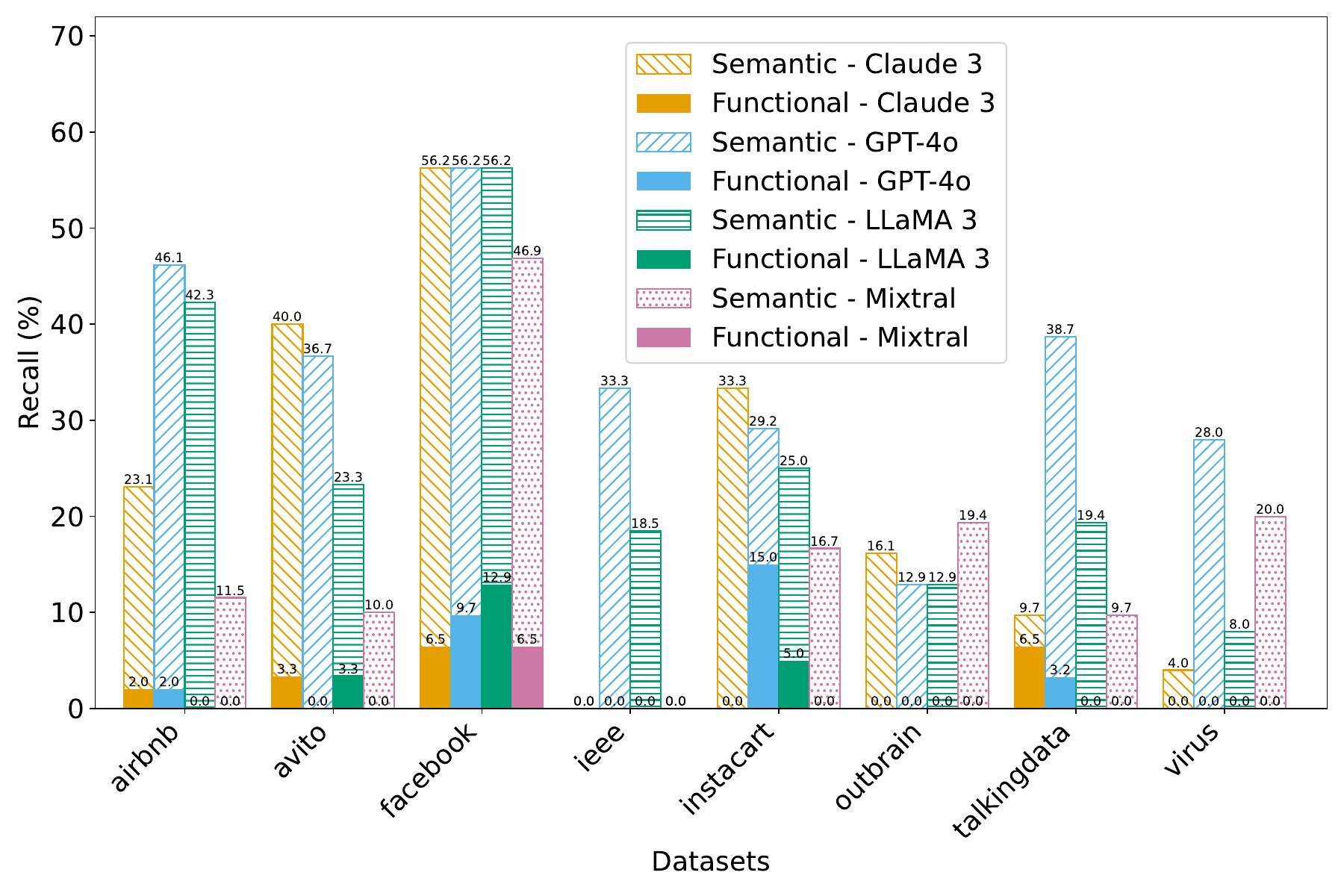}
    \caption{Recall metrics of different LLMs hitting golden features measured by semantic and functional similarity.}
    \label{fig:equivalence}
    \vspace*{-3mm}
\end{figure}

Figure~\ref{fig:equivalence} shows the alignment results between LLM-engineered features and golden features, measured by recall metrics using both $M_{sem}$ and $M_{func}$. Overall, the recall percentages are higher for $M_{sem}$ compared to $M_{func}$, indicating that while LLMs can generate features that are semantically similar to the golden features, they struggle to ensure these features are functionally equivalent. GPT generally leads in both metrics, highlighting its effectiveness in feature generation.

\begin{table}
  \caption{Downstream model performance with different features. Top-1 and top-2 scores are bold-faced and underlined respectively.}
  \label{tab:results-with-data}
  \small
  \begin{tabular}{ccccc}
    \toprule
    \textbf{Features} & \textbf{Avito} & \textbf{IEEE-CIS} & \textbf{Outbrain} & \textbf{Instacart} \\
    \textbf{Metric} & LogLoss $\downarrow$ & AUC $\uparrow$ & AUC $\uparrow$ & AUC $\uparrow$ \\
    \midrule
    \textbf{Raw Feat.} & 0.0337 & 0.9108 & 0.5321 & 0.4977 \\
    \midrule
    \textbf{GPT-4o} & 0.0355 & 0.9210 & 0.5194 & 0.7620 \\
    \textbf{Claude 3} & \underline{0.0328} & 0.9100 & 0.5298 & \underline{0.8437}  \\
    \textbf{LLaMA 3} & \textbf{0.0321} & 0.9058 & \underline{0.6013} & 0.5000  \\
    \textbf{Mixtral} & 0.0340 & \underline{0.9213} & 0.5971 & 0.5345  \\
    \midrule
    \textbf{Golden Feat.} & 0.0331 & \textbf{0.9236} & \textbf{0.6164} & \textbf{0.8526}  \\
  \bottomrule
\end{tabular}
\vspace*{-3mm}
\end{table}

For the Facebook dataset, all models perform well semantically (56.25\%) and functionally, with LLaMA3 leading at 12.90\%, followed by GPT-4o (9.68\%). This indicates that the features required for Facebook are easier for LLMs to generate both semantically and functionally. Conversely, datasets like IEEE-CIS, Outbrain, and Virus present significant challenges, with low or zero recall scores across all models for both metrics. Notable golden features in these datasets include frequency counting, feature interaction, and features grouped by multiple columns, all of which are difficult for existing LLMs (see Sec.~\ref{subsec:rq2} for deeper analysis). Additionally, IEEE-CIS has particularly strong features created by grouping by the ``card1'' ID column, but LLMs tend to group by other columns. No single model consistently outperforms across all datasets. While GPT-4o shows strong overall performance, Claude3 excels in specific cases such as Avito, with a 40\% recall for $M_{sem}$ but struggles with $M_{func}$. In conclusion, these recall comparisons highlight areas where LLMs may be improvable, particularly in generating functionally equivalent features.

Table~\ref{tab:results-with-data} compares the performance of models using LLM-generated features with those using golden features and raw data alone. The performance metrics vary by dataset, with LogLoss used for Avito and AUCROC used for IEEE-CIS, OutBrain, and Instacart. We found that golden features generally lead to the best model performance across these datasets, demonstrating the superior quality of expert-crafted features. LLM-generated features are generally useful, as they consistently improve model performance over using raw features alone. Although GPT-4o performs well in feature recall metrics, its downstream model performance surprisingly lags behind. This discrepancy is likely because GPT-4o tends to generate more features than other models, which can introduce more noise and complexity, leading to overfitting and reduced generalization. For instance, in the Instacart dataset, GPT-4o generates 35 features but achieves an AUCROC of 0.7620, whereas Claude3, with only 11 features, achieves a higher AUCROC of 0.8437. Notably, LLaMA3 and Claude3 show potential in outperforming human baselines in specific cases. For example, in the Avito dataset, Claude3 (0.0328) and LLaMA3 (0.0321) achieve better LogLoss than the human-engineered features (0.0331), indicating that LLM-generated features can sometimes surpass expert-crafted ones, especially when they effectively balance feature quantity and quality.

\subsection{RQ2: Patterns LLMs Excel or Struggle with}\label{subsec:rq2}

% \begin{table*}
%   \caption{Feature categorizations}
%   \label{tab:feature-categorizations}
%   \begin{tabular}{p{0.1\linewidth} p{0.3\linewidth} p{0.5\linewidth}}
%     \toprule
%     Category & Description & Example Pandas code template \\
%     \midrule
%     Unary transforms & Compute a unary function of a column & \texttt{pd.to\_datetime(A['X'])} \\
%     N-ary transforms & Compute a function involving multiple columns & \texttt{A['X'] + A['Y']} \\
%     Simple joins & Merge columns from another table based on a key & \texttt{A.merge(B, on='X', how='left')} \\
%     Single-key groupby's & Compute statistics of columns grouped by a single key & \texttt{A = A.join(B.groupby('X')['Y'].mean(), on='X', how='left'} \\
%     Multi-key groupby's & Compute statistics of columns grouped by a combination of keys & \texttt{A.groupby(['X', 'W'])['Y'].mean()} \\
%     Nested groupby's & Perform groupby operations within other groupby operations to calculate statistics & \texttt{A.groupby('X').apply(lambda g: g.groupby('Y').sum())} \\
%     Temporal groupby's & Compute statistics of columns grouped by some key, but only over rows whose timestamp columns are before its own timestamp & \texttt{A.groupby('X').rolling('7d', on='T')['Y'].sum()} \\
%     Non-builtin aggregations & Use custom aggregation functions to compute statistics. & \texttt{A.groupby('X').agg({'Y': lambda x: x.quantile(0.9)})} \\
%     Time-lagged features & Compute features based on time differences between events (rows) & \\
%     \bottomrule
%   \end{tabular}
% \end{table*}

To study which kind of features LLMs are better generating, we further break down the golden features into two categories:

\begin{itemize}[leftmargin=1em,topsep=0pt]
  \item \textbf{Feature transforms}, including
  \begin{itemize}[leftmargin=1em,topsep=0pt]
  \item \textbf{Unary transforms}: a unary function of a column.
  \item \textbf{N-ary transforms}: a function involving multiple columns.
  \item \textbf{Time-lagged features}: features based on time differences between events (rows), e.g. "finding the difference between the number of calls from the same each day with respect to the day before".
  \end{itemize}
  \item \textbf{Joins and aggregations}, including
  \begin{itemize}[leftmargin=1em,topsep=0pt]
  \item \textbf{Simple joins}: columns merged from another table.
  \item \textbf{Single-column group-by}: statistics of columns grouped by a single column.
  \item \textbf{Multi-column group-by}: statistics of columns grouped by a combination of columns.
  \item \textbf{Nested group-by}: statistics from group-by operations within another group-by operations, e.g. "finding the minimum average of call durations per device over all mobile devices for each user".
  \item \textbf{Temporal group-by}: statistics of columns grouped by some column, but only over rows whose timestamp columns are before its own timestamp, e.g. "counting the number of calls from the same device before the current call".
  \item \textbf{Non-builtin aggregations}: statistics with custom aggregation functions unavailable in Pandas, e.g. entropy.
  \end{itemize}
\end{itemize}

\begin{figure}
    \centering
    \includegraphics[width=0.6\linewidth]{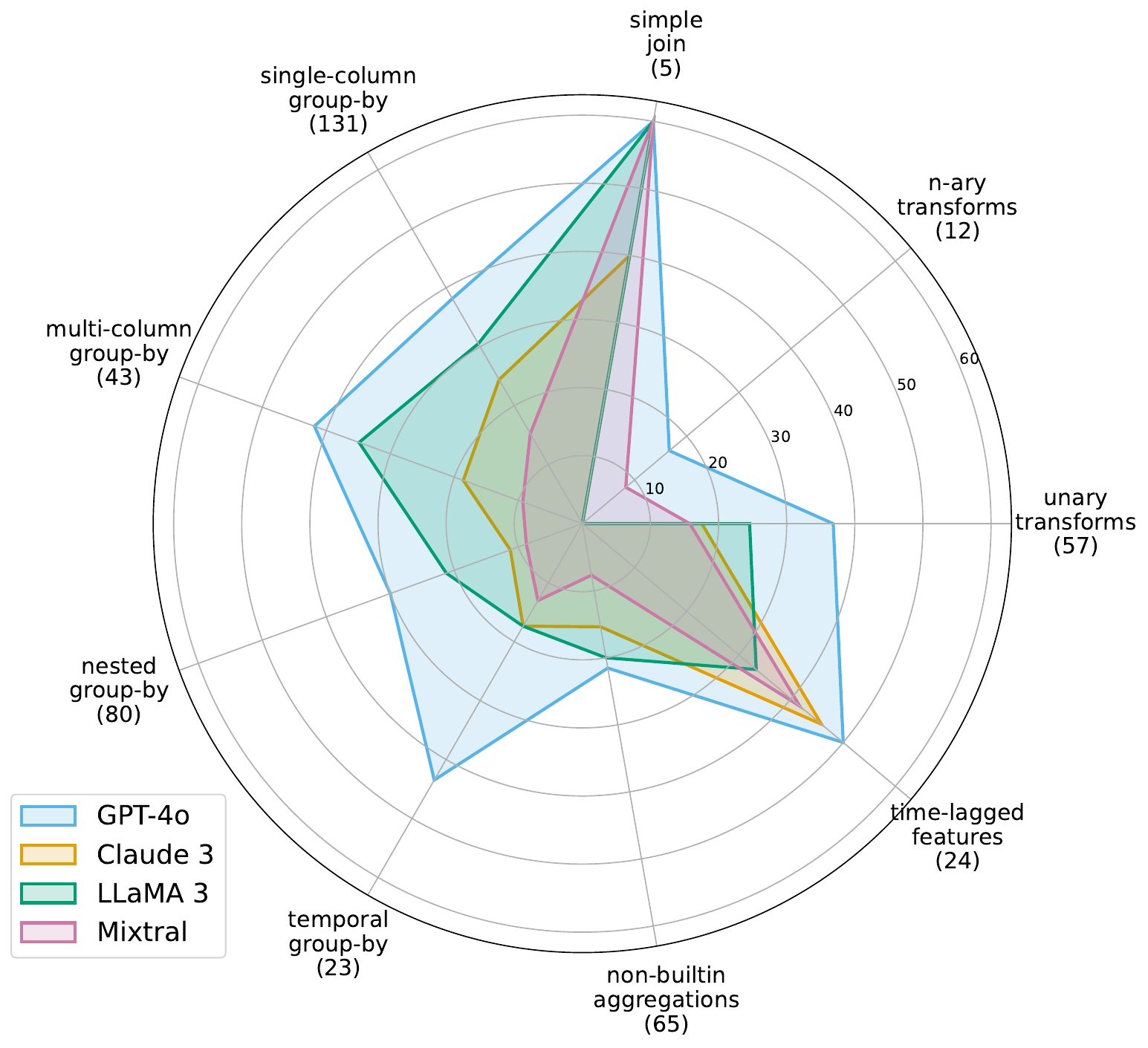}
    \caption{Recall scores (measured by semantic similarity) grouped by feature category. The numbers in parentheses represent the number of golden features within that category.}
    \label{fig:category-breakdown}
\vspace*{-6mm}
\end{figure}

The statistics of such features, as well as the category-wise recall by semantic similarity, are shown in Figure~\ref{fig:category-breakdown}.  A majority of golden features are \textbf{joins and aggregations}.  This is expected as the downstream models contestants used, such as gradient-boosted decision trees \cite{ke2017lightgbm, chen2016xgboost} or factorization machines \cite{rendle2010factorization,juan2016field}, inherently incorporate feature transformations and interactions to some extent.  Notably, GPT-4o outperformed the other LLMs in all categories.  However, none of the LLMs achieved 50\% recall in any categories, with the exception of \textbf{time-lagged features} and \textbf{simple joins} --- categories that are simpler and have fewer golden features (5 and 24, respectively).  The figure also highlights that LLMs struggle in generating complex features like \textbf{n-ary transforms}, \textbf{nested group-by} and \textbf{non-builtin aggregations}.  These categories require designing and implementing highly specialized functions that often demand deep domain-specific knowledge, which goes beyond general common sense.

\section{Conclusion}

This paper proposes \sys, a framework for evaluating the feature engineering capability of LLMs, by curating a dataset with 251 human-engineered features from 8 Kaggle competitions.  Initial experiments on 4 LLMs reveal varying gaps relative to human-level feature engineering expertise; in particular, we observe that while LLMs in may excel generating simple features, they struggle at times to generate features involving complex functions, aggregations, and table joins.  We believe that addressing these challenges will be crucial for realizing the full potential of LLMs in automating data science tasks.

\bibliographystyle{ACM-Reference-Format}
\balance
\bibliography{main}

\appendix

\end{document}